# Speeding up design and making to reduce time-to-project and time-to-market: an AI-Enhanced approach in engineering education[*]

Giovanni Adorni[1†]  Daniele Grosso[2‡]

[1] DIBRIS - University of Genoa - giovanni.adorni@unige.it

[2] DIFI - University of Genoa - daniele.grosso@unige.it

**Abstract**

This paper explores the integration of AI tools, such as ChatGPT and GitHub Copilot, in the Software Architecture for Embedded Systems course. AI-supported workflows enabled students to rapidly prototype complex projects, emphasizing real-world applications like SLAM robotics. Results demon-started enhanced problem-solving, faster development, and more sophisticated outcomes, with AI augmenting but not replacing human decision-making.

**Keywords**

Human-AI Collaboration, Prompt-based Learning, AI-Supported Design and Development

## 1. Introduction

The integration of Artificial Intelligence (AI) has become essential in educating engineers, scientists, and graduates. This paper discusses experiences from the "Software Architecture for Embedded Systems" of the Master in Mechatronics Engineering at the University of Genoa (https://corsi.unige.it/off.f/2024/ins/79120?codcla=9269), introducing an AI-assisted teaching approach.

The course was divided into two phases:

1. Students acquired skills in designing and developing embedded systems, learning concepts such as IoT, human-machine interfaces, architecture design, software programming, and UML diagrams.
2. Generative AI was introduced as a co-designer to assist throughout the development process, supporting design solutions, generating documentation, and aiding in software development and testing.

This structure aligns with key principles of active learning and constructivist pedagogy, where students engage with real-world problems, apply theoretical knowledge, and make decisions based on critical thinking [1, 2]. AI scaffolds the learning process, allowing quicker iteration on designs.

A key innovation is *prompt-based learning*, where students interact with AI systems through carefully designed prompts. This technique increases student engagement and teaches effective AI communication. Research shows prompt-based learning can be highly effective in developing problem-solving skills, especially when students iteratively refine their prompts [4, 5].

The use of AI in coding, debugging, and documentation supports the pedagogical goal of reducing cognitive load. By automating routine tasks, AI allows students to focus on higher-order thinking, such as refining system architecture and solving complex design problems. This aligns with the concept of distributed cognition, where cognitive processes are shared between the student and AI, enhancing the learning experience [6].

---

Motivations for introducing Generative AI include:
- Design Optimization: Students use AI to efficiently explore multiple design options, mirroring real-world engineering practices [7, 8].
- Automated Documentation: AI generates UML diagrams and project documentation, allowing students to focus on iterative design aspects.
- Coding and Testing Assistance: Generative AI assists with coding and debugging, providing immediate feedback and automating many testing aspects.

This approach reflects a broader trend: the democratization of coding. Google CEO Sundar Pichai noted that AI is reshaping programming, making it more accessible to a broader audience [9]. This democratization is evident in the course, where students with varying programming expertise can build complex systems [10].

Recent studies confirm that integrating AI into educational contexts can significantly improve learning outcomes. Research on AI-assisted design and development shows students can produce higher quality projects in less time [7, 11].

## 2. AI Integration in Education

Generative AI tools such as ChatGPT (https://openai.com/), Claude (https://claude.ai/), GitHub Copilot (https://github.com/features/copilot), and LM Studio (https://lmstudio.ai/) were introduced to assist students throughout the project lifecycle, including design, documentation, coding, debugging, and final reporting. These tools played a pivotal role in helping students navigate complex project requirements.

AI integration began during the design phase, where tools facilitated rapid generation of design alternatives. Students prompted the AI to explore different architectural options, mirroring industry practices. GitHub Copilot supported coding by generating suggestions for code snippets and helping to automate repetitive tasks.

A unique element of the course was the use of AI to generate project documentation, particularly UML diagrams. AI tools helped create accurate representations of system architectures, allowing students to focus on refining the design rather than manual diagramming. LM Studio provided locally installed AI tools to support different programming environments.

Students engaged in prompt-based learning, carefully constructing and refining prompts to elicit the most relevant and accurate results from AI systems [3]. This interaction not only improved problem-solving skills but also taught the importance of clear communication with AI tools.

Students were tasked with continuously refining their prompts based on AI feedback, resulting in more sophisticated project outcomes. For example, students working on robotics projects could ask the AI to generate specific code segments for telemetry systems or suggest optimizations for Simultaneous Localization and Mapping (SLAM) algorithms.

An innovative aspect of this course was the requirement to document AI interactions. Students saved copies of their prompts and AI responses, providing a detailed record of AI's influence on project development. This documentation served as a reflective tool, allowing students to assess their AI use effectiveness and identify areas for improvement in prompt strategies.

The documentation process was pedagogically significant as it allowed students to reflect on the collaborative nature of their work with AI, identifying key break through moments and instances where human judgment was necessary to override AI suggestions. This reflective practice aligns with educational theories emphasizing metacognition as a critical element of deep learning [12].

While all student groups used AI tools, the extent and manner of their use varied. Some teams primarily used AI to generate and debug code, while others focused on using AI to generate UML diagrams and optimize design workflows. These variations highlighted the flexibility of AI tools to adapt to different project needs and team dynamics.

A notable example was the use of AI in SLAM-based robotics projects. AI tools helped students rapidly prototype systems capable of telemetry and mapping, significantly reducing development time. Students developed fully functional prototypes within the course's tight timeframe, a challenging task without AI assistance. Although AI had advanced capabilities, students remained the final decision-makers in every aspect of their projects, ensuring AI functioned as a supportive tool rather than a replacement for critical human input.

## 3. Evaluation and Prompt-Based Learning in AI-Supported Project Development

The assessment process in the "Software Architecture for Embedded Systems" course was designed to mirror real-world project development, with students working in teams of three. Each group was tasked with developing a functional prototype supported by comprehensive project documentation, simulating different roles within a project team. This collaborative approach allowed students to experience different responsibilities such as project management, coding, and testing. The evaluation focused on the delivery of a working prototype and the submission of a detailed project report, including a log of all AI prompts used during development.

**Project-Based Assessment Structure** - Students selected projects aligned with their interests, collaborating with faculty to define requirements. This phase focused on understanding the project's scope, technical challenges, and deliverables. During development, students utilized AI tools for project management, documentation generation, coding, and debugging. They were required to critically evaluate AI output, ensuring that final decisions were based on their own analysis and understanding. This exercise reinforced the human-AI collaboration dynamic: AI as a powerful assistant, with students retaining responsibility for project completion.

**The Role of Prompt-Based Learning** - A critical aspect of the evaluation was the use of AI-driven prompts. Students submitted a comprehensive list of prompts used during development, demonstrating their ability to interact with and guide AI systems. The prompt log was evaluated based on how iterative and refined the prompts became over time. This aspect tested both technical proficiency and students' ability to engage in metacognitive reflection, considering how their interactions with AI impacted project progress.

**Project Presentation and Evaluation** - The exam culminated in a dual format presentation:
1. Non-Technical Presentation: Students presented their project to a non-technical audience, simulating scenarios where engineers must communicate complex ideas to management or customers.
2. Technical Presentation: Students provided a detailed explanation of technical challenges faced, solutions implemented, and rationale behind their design decisions. They discussed how AI assisted in project development and where human intervention was necessary.

Both aspects of the presentation were essential to the evaluation process, testing students' ability to communicate effectively to diverse audiences and demonstrate depth of understanding, problem-solving skills, and mastery of both technical concepts and AI collaboration.

Overall, students were evaluated on:
- Prototype functionality
- Documentation quality
- Effective use of AI tools while maintaining project control

This comprehensive approach ensured that students developed technical skills and learned to critically evaluate and collaborate with AI tools, preparing them for real-world engineering challenges.

## 4. Example Project: SLAM-Capable Robot Development

Among the proposed projects, two different groups of students focused on developing a robot with SLAM (Simultaneous Localization and Mapping) capabilities. The selected projects involved integrating various hardware components, including an Arduino Mega, a Raspberry Pi, ultrasonic sensors, motor drivers, and encoders, as well as a custom software architecture designed to manage the robot's navigation.

As an example, we report results from one of the five projects assigned this academic year, which focused on developing a robot capable of sampling parameters from the environment and using them to generate a map and locate its position (see Figure 1).

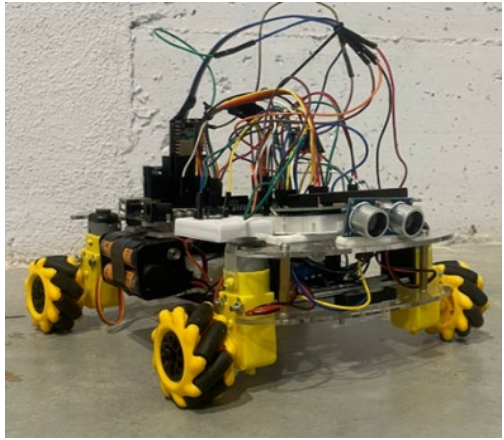

**Figure 1.** Robot prototype with omnidirectional wheels developed by a group of students.

Their project stood out for its innovative approach to combining AI decision-making with physical robot control. The main achievements and challenges of this project were:

- Hardware setup - The group was tasked with sending telemetry data collected by the robot to either a desktop computer or a Raspberry Pi, and sending images taken periodically by an on-board camera. This setup allows for further development and implementation of machine learning algorithms for object recognition and real-time navigation decision-making, utilizing the computing power available on external computers.
- Sensor Fusion - The project involved implementing algorithms to combine data from multiple sensors for improved environmental awareness and navigation accuracy, enhancing the precision of the robot's position estimation.

The teams developed an advanced omni-directional wheel robotic system with adaptive control for environmental learning, cloud integration for complex AI processing, and human-robot interaction capabilities including remote control and basic speech recognition. They successfully overcame challenges related to real-time processing, power management, and software integration, resulting in a highly capable and efficient robotic platform.

These students demonstrated exceptional project management, technical implementation, and presentation skills. Their successful completion of the projects and excellent performance on the exams highlighted the effectiveness of the course structure in preparing students for real-world engineering challenges.

The project documentation and technical details about each aspect of the project discussed during the exam include management, system architecture, coding, and documentation. The group documented the prompts used to generate code snippets, UML diagrams, and debugging suggestions.

### 4.1. Project management

Every aspect of project management is discussed during the exam; WBS, Gantt and Pert diagrams, resources and their costs, budget and any related issues. Let's look at an example of an AI assisted WBS generation prompt using PlantUML (https://plantuml.com/) as a tool that facilitates the quick and easy creation of a wide range of diagrams:

> *Give me PlantUML code for a WBS based on what I send you.*

The response was:

*"Certainly, I can help you create a WBS (Work Breakdown Structure) diagram using PlantUML. Please send me the details of your project, including the various levels and activities you want to represent."*

An iteration is required here; once the material was provided, the response to the prompt was as follows:

*"To create a WBS (Work Breakdown Structure) using PlantUML, we can divide the project into its main components and sub-activities. Here's an example of how you could structure your project using PlantUML."*

Due to the complexity and size of the diagram, the materials are not included, but the system commentary is given below:

*"This code generates a WBS structure for your project. If you wish to include more detailed information such as dates, durations, or names of people involved, you can further customize the nodes, although PlantUML for WBS is primarily used to represent the hierarchical structure of activities rather than the specific details of the project."*

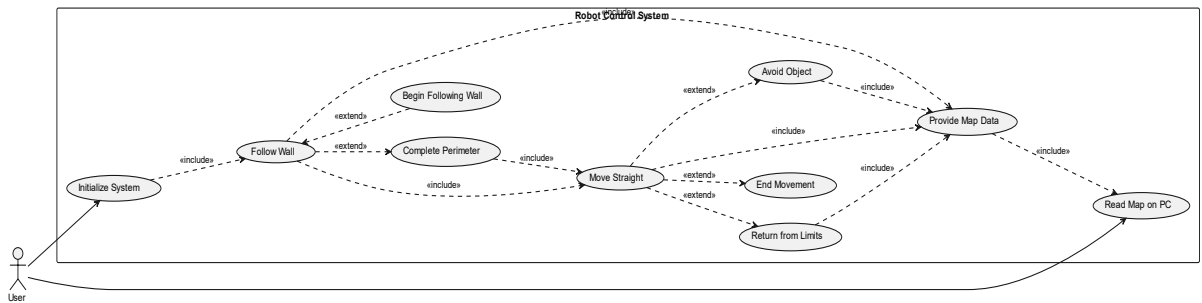

**Figure 2.** Example of a Use Case diagram.

### 4.2. System Architecture

This aspect also requires UML diagrams that illustrate the robot's components and their interactions, generated with AI support. The diagrams outline subsystems such as sensor integration, motor control, and data transmission.

As part of the course, students learned to use AI to generate UML diagrams, a critical skill in software documentation. Below is an example of how AI was used to create UML diagrams: Use Case, State Diagram, and Class.

The Use Case diagram is a visual representation that shows how users interact with a system to achieve specific goals. It highlights the different ways in which users can engage with the system and illustrates the different interactions between the users (actors) and the system itself.

Below is an example of a prompt for creating a Use Case diagram:

*> Create a PlantUML use case diagram for a Robot Control System with the following use cases: - Initialize System - Follow Wall - Move Straight - Avoid Object - Return from Limits - End Movement - Read Map on PC Include a User actor. Show relationships between use cases, such as 'Begin Following Wall' extending from 'Follow Wall' to 'Move Straight'. Use 'include' relationships for repeated behaviors like 'Provide Map Data'.*

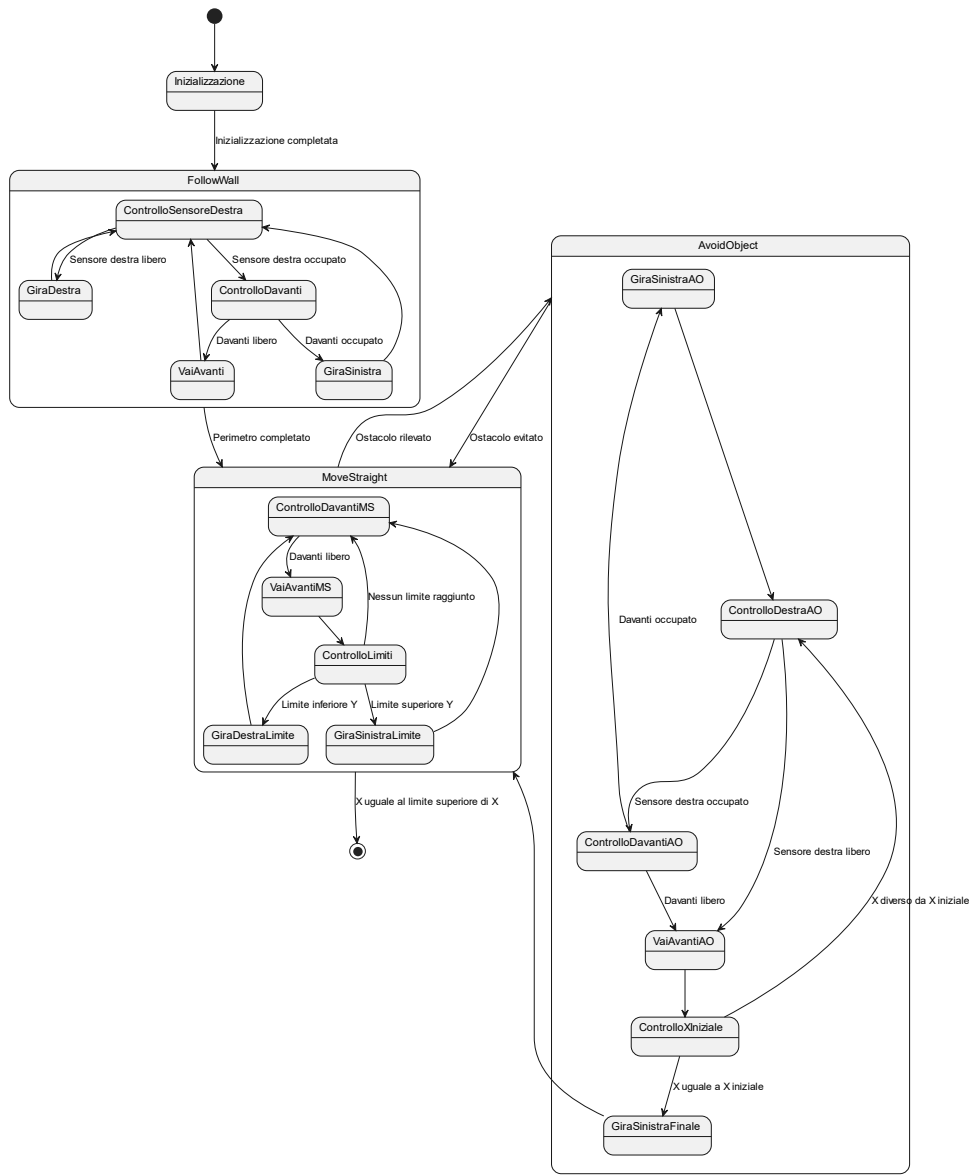

**Figure 3**. Example of a State Chart diagram.

Once the plantUML code is generated, it can be pasted and eventually refined to produce an image ready for the documentation files, as shown in Figure 2.

The State Chart diagram, on the other hand, focuses on the dynamic behavior of a system. It illustrates the different states an object can be in and the transitions between those states. This diagram is particularly useful for understanding how a system responds to various events and changes over time.

The following is an example of a prompt for creating a State Chart diagram:

> *Generate a PlantUML state diagram for a robot's movement control system with the following main states:*
> *- MoveStraight*
> *- AvoidObject*
> *- FollowWall*
> *- Inizialization*
> *Include substates and transitions, especially for MoveStraight (e.g., ControlloDavantiMS, VaiAvantiMS) and AvoidObject (e.g., GiraSinistraAO, ControlloDavantiAO). Show conditions for transitions like 'Davanti libero' and 'Davanti occupato'. Include a final state and an initial state.*

The response allows for the generation of the diagram shown in Figure 3.

Finally, the class diagram provides a structural view of a system. It depicts the classes of the system, along with their attributes and methods, and shows the relationships between these classes. This diagram is essential for modeling the static aspects of a system, providing a clear picture of its overall architecture. As shown above, a well-formed user prompt allows you to generate what the designer intended, and it is also possible to refine it.

The following is an example of a prompt for generating the class diagram:

> Create a PlantUML class diagram for a robot control system with the following classes: powerUnit, Motore, Encoder, PIDController, Giro, ultrasonico, ConnessioneWiFi, ConnessioneMQTT, PubSubClient. Include important attributes and methods for each class. Show relationships between classes, especially how the 'powerUnit class' interacts with others such as 'Motore' and 'Encoder'.

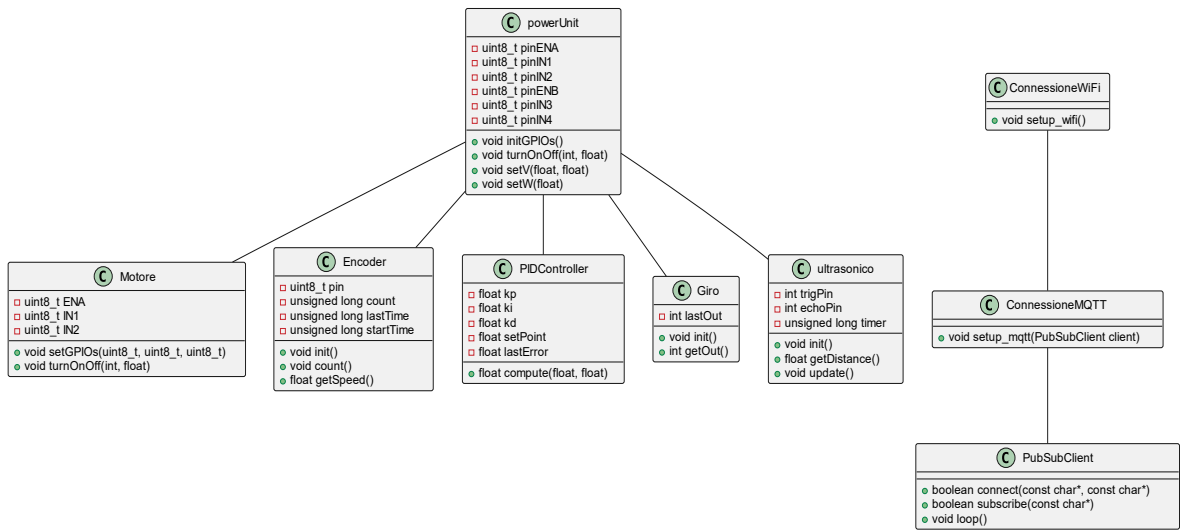

**Figure 4.** Example of a Class diagram.

The response allows the appropriate diagram to be generated, as shown in Figure 4. Note that all the prompts must be very precise, and this requires an iterative process.

### 4.3. Code and Debugging Log

The students provided a complete codebase developed using GitHub Copilot, with sections of code where AI-assisted debugging occurred. They provided a technical report detailing algorithmic challenges faced, including inconsistent sensor data and motor control issues, and how AI helped overcome these challenges by generating optimized solutions. Here's an example of user-AI interaction:

> Write me the code you use for the initialization of powerUnit and let's see if we can identify the problem. I'll give you the .h file first, then the .cpp file.

The answer:

"Perfect, share the code of the .h file and then the .cpp file that you use for the definition of the powerUnit class"

Once provided the necessary input:

"Thank you for sharing the files. Let's examine the problem:
1. Initialization of powerUnit: Your declaration and definition seem correct. However, if powerUnit is not initializing correctly, there are a few things we can check.
2. Constructor verification: Ensure that all pins passed to the constructor are valid and there are no pin conflicts.

3. *Debugging with Serial: Add debug messages inside the constructor of the powerUnit class to verify if it's being called correctly.*
4. *Constructor call: Make sure that the call PU = new powerUnit(...); in the main file is correct.*
*Here are some modifications you can make for debugging."*

The following are instructions and fixes to also improve logging to find the source of the problem, considering both hardware and software aspects:

*"Make sure that all pins are correctly connected and there are no conflicts. ..."*

This process repeated several times, also by providing the AI with the output to be interpreted and asking it to identify the causes of the error and to eliminate them, makes it possible to speed up the development process; the usual mental schemas are followed, but one takes advantage of the AI's.

The WBS and UML examples show how AI can help with project planning and system design, while the code generation and debugging examples demonstrate the potential of AI to speed up development cycles.

This approach allowed students to focus on higher-level problem solving and system design, while using AI to handle more routine tasks. The experience of working with AI tools prepared students for the evolving landscape of robotics and embedded systems engineering, where AI-driven development is becoming more prevalent. This unique blend of hands-on experience and AI integration has provided students with valuable skills essential for future careers in technology-driven fields.

## 5. Conclusions, Final Considerations, and Future Developments

This first attempt at integrating AI into the Software Architecture for Embedded Systems course proved highly successful, with five student groups delivering projects of exceptional quality. The use of AI throughout the project lifecycle - particularly in coding, debugging, and documentation - allowed students to focus on critical problem solving, resulting in more complex and innovative projects than in previous years. AI facilitated rapid iteration and effective decision-making, particularly in hardware-software integration, real-time data processing, and machine learning implementations in constrained environments.

The students' ability to adapt AI tools to their specific project needs demonstrates the power of prompt-based learning and AI-assisted workflows in modern engineering education. Each group successfully completed their project within the given deadline (one month), and all received high marks give their evident skills, highlighting the value of AI as a co-designer in development.

### 5.1. Reflections and Future Directions

All eight students in this year's course expressed strong satisfaction with the AI-integrated approach and recommended its continuation. While formal statistical comparison was limited by sample size, the two project groups completed approximately twice the workload of previous years' groups in the same timeframe, demonstrating significant efficiency gains.

While AI significantly reduced the technical challenges students faced, some conceptual difficulties remained, particularly in areas that required deeper technical understanding. AI was instrumental in many tasks, but it could not replace the hu-man decision-making process; thus students need to strengthen their foundational knowledge.

In the 2024-2025 academic year, we aim to expand the course, increase the number of participating groups, and improve the integration of AI tools to ensure that students develop skills aligned with the evolving needs of industry. A key focus will be to improve students' understanding of AI prompting techniques to maximize their interactions with AI systems. In addition, we will explore the use of local AI systems to address privacy and security concerns that are critical in industrial applications.

We also aim to introduce more hands-on resources, such as a 3D printing lab and other tools, coupled with AI support.

**5.2. Broader implications and applications**

The success of AI integration in this course has broader implications for other disciplines. We see great potential for the application of Generative AI in fields such as applied physics, medicine, and environmental engineering, where complex problem solving, and data processing are essential. AI is ready to revolutionize education in many disciplines, and its role in embedded systems is just the beginning.

   In conclusion, the first iteration of AI-enhanced education in this course proved to be both effective and impactful, even considering that all projects were completed by students in one month and showed an high quality level even if they required many advanced skills. We believe that continued refinement of these methods will prepare students to engage with AI technologies in their future careers and equip them with the skills necessary to navigate the evolving landscape of engineering and technology.

   Future course developments will focus on strengthening prompt engineering skills and establishing structured methodologies for AI-assisted development. Priva-cy concerns will be addressed through specialized prompting strategies and locally installed open-source AI tools like LM Studio, ensuring secure handling of sensitive data.